# Robots Understanding Contextual Information in Human-Centered Environments using Weakly Supervised Mask Data Distillation

Daniel Dworakowski, and Goldie Nejat

*Abstract*—Contextual information contained within human environments, such as signs, symbols, and objects provide important information for robots to use for exploration and navigation. To identify and segment contextual information from complex images obtained in these environments data-driven methods such as Convolutional Neural Networks (CNNs) can be used. However, these methods require significant amounts of human labeled data which can be very slow and time-consuming to obtain. Weakly supervised methods can be used to address this limitation by generating pseudo segmentation labels (PSLs). In this paper, we present the novel Weakly Supervised Mask Data Distillation (WeSuperMaDD) architecture for autonomously generating PSLs using CNNs not specifically trained for the task of context segmentation, e.g. CNNs alternatively trained for: object classification (dog or cat), image captioning (the dog is sitting on the couch), etc. WeSuperMaDD is uniquely able to generate pseudo labels using learned image features (e.g. edges, contours, etc.) from sparse data and data with limited diversity, which are common in robot navigation tasks in human-centred environments (i.e. shopping malls, grocery stores, etc.). Our proposed architecture uses a new mask refinement system which automatically searches for the PSL with the fewest foreground pixels that satisfies cost constraints measured by a cost function. This removes the need for handcrafted heuristic rules. Extensive experiments were conducted to successfully validate the performance of our WeSuperMaDD approach in generating PSLs for datasets containing text of various scales, fonts, and perspectives in multiple varying indoor/outdoor environments (e.g. public transit vehicles, roads, etc.). A detailed comparison study conducted with existing Naive, GrabCut, and Pyramid approaches found a significant improvement in label and segmentation quality. Furthermore, a context segmentation CNN trained using the WeSuperMaDD architecture achieved measurable improvements in accuracy when compared to a context segmentation CNN trained with Naive PSLs. We also found our method to have comparable performance to existing state-of-the-art one-stage and two-stage text detection and segmentation methods on real text datasets without requiring any segmentation labels for training.

## I. INTRODUCTION

HUMAN-centered environments contain an abundance of contextual information such as signs, symbols, objects, and text that are used as landmarks to help guide users with point-to-point navigation in unknown environments [1], and update maps of the environment [2]. Service robots working in varying human-centered environments can exploit these types of contextual information to aid with navigation. For example, robots can use aisle signs in grocery stores to determine which aisles to search for a particular item [3]. They have also used contextual information for mapping and localization. Namely by using an annotated map of an office with room placards for goal directed navigation [4]. Robots have also created semantic maps using product locations [5], maps from unique text landmarks identified in images [6], and have used salient objects identified from learned features (e.g. edges, contours, etc.) for visual odometry [7]. These approaches rely specifically on a robot's ability to identify and localize context in an environment.

Recent work in the area of context detection has made use of Convolutional Neural Networks (CNNs) to detect the presence of various types of objects within images of an environment [8], [9]. However, the amount of human effort required to generate the vast expert labeled datasets required for training these existing networks significantly increases the overall time cost of their use. While datasets such as the Waymo open dataset [10], Open Images [11], and ICDAR-15,17 [12], [13] contain several context categories that can be used by robots, e.g. text, cars, etc., they are not fully labeled, requiring the classification of each pixel in an image in the dataset prior to training. Creating labels for all context instances within these datasets is significantly time consuming and must be done manually. Past research has found that the time required to manually generate a segmentation label is approximately 54s per context instance [14]. Based on this we can estimate that it would take 20 years to segment all 2D objects in the Waymo open dataset!

Given the large investment needed to create these labels, several automated methods have been proposed to generate pseudo segmentation labels (PSLs) to segment an input image and to replace human expert labels. In general, naive methods have been proposed to avoid the human time-cost of per-pixel segmentation by: 1) using bounding box labels [15], or 2) generating soft labels (i.e. labels between 0 and 1) based on the assumed shape of context instances [16]. However, these approaches rely on specific assumptions about the shape and structure of context instances that are not always valid, for example, assuming objects do not have holes (e.g. the center of the character 'o').

Recently, semi or weakly supervised learning methods have been proposed to introduce learned features into the label generation process. Semi-supervised methods train a model using a dataset containing fully labeled and unlabeled data to generate pseudo labels for the unlabeled subset [17]. An example of semi-supervised learning is data distillation where the fully labeled subset of a dataset is used to train a CNN. The CNN is then used to generate labels for unlabeled images using an ensemble of predictions from multiple transformed versions of each image [8]. These methods, however, still require human expert supervision to generate the labeled set. Alternatively,

This work was supported by AGE-WELL Inc., the Natural Sciences and Engineering Research Council of Canada (NSERC), the Canada Research Chairs program (CRC), the Vector Institute Scholarship in Artificial Intelligence, and the NVIDIA GPU grant.

The authors are with the Autonomous Systems and Biomechatronics Laboratory in the Department of Mechanical and Industrial Engineering, University of Toronto, 5 King's College Road, Toronto, ON, M5S 3G8 Canada. (e-mail: daniel.dworakowski@mail.utoronto.ca; nejat@mie.utoronto.ca).

weakly supervised methods generate PSLs using partial information such as class or bounding box labels to completely label a dataset [18]. For example, weakly supervised methods use techniques such as class peak response [19], adversarial erasing [20], and Class Activation Maps (CAMs) [21]. However, these approaches can significantly restrict the types of CNNs that can be trained due to the introduction of specialized CNN layers. These layers are used to determine the contribution of pixels in the input image to the CNN's final output. For example, CAMs require an additional global average pooling layer followed by a fully connected layer. Requiring specific layers limits the applicability of such existing weakly supervised segmentation methods since they cannot be generalized to all problems.

To avoid fully training a network, in [15], unsupervised segmentation methods, such as GrabCut [22] were used. However, since this method is class agnostic, it results in significant noise being present in the generated labels, it reduces the performance of the trained models compared to fully supervised models [15].

As robots must operate in different environments with varying terrain, configurations, and objects, large datasets must be obtained to train a robust segmentation CNN for context detection. The size of these datasets makes generating segmentation labels time consuming and slow. Using weakly supervised methods to generate PSLs is desirable as they do not require fully labeled data and can therefore reduce manual labeling effort. However, weakly supervised methods require training with large diverse datasets to learn a feature representation that must then be segmented using handcrafted heuristic rules. Moreover, training must be repeated if new data is added to a dataset [19]–[21] thus increasing the computational cost of their use. Alternatively, we propose using readily available CNNs that are not trained specifically for segmentation (but rather object classification, image sequence prediction, etc.) as their convolutional layers can readily be transferred to the segmentation task [23]. These CNNs are trained with sparse labels, such as the overall class of an object (chair or desk), or the classes of a sequence of objects (cat and dog), rather than per-pixel image classification (which exact pixels in an image contain a dog). In order to segment contextual information using PSLs, features within these classes (e.g. edges, contours) need to be identified. By extracting the features used by these CNNs and incorporating them into a weakly supervised learning framework we minimize the need for additional training, while providing a source of informative image features to guide label generation.

In this paper, we present a novel autonomous Weakly Supervised Mask Data Distillation pseudo label generation architecture, WeSuperMaDD, for the segmentation of contextual information in varying environments using a partially labeled dataset containing bounding boxes and context labels (e.g. object class labels, image caption labels, etc.). These labels can be used to directly train existing CNN models for robotics tasks such as context identification and recognition, including for robot navigation, exploration, and obstacle avoidance. Our main contributions are: 1) we present the first architecture which uses learned features extracted from existing CNNs not trained specifically for segmentation in order to generate PSLs regardless of dataset size or diversity, and 2) we introduce an autonomous mask refinement system that uniquely searches for PSLs that both have the fewest foreground pixels and satisfy cost constraints as measured by a cost function. By combining these learned features and our new mask refinement process, we remove the costly need for human experts to train a CNN for PSL generation and to create the handcrafted rules for segmentation that are typically required in previous works.

The main advantage of our architecture is that it allows for the robust generation of PSLs in partially labeled datasets obtained by robots as they explore their environments, without the need for large diverse datasets. The proposed method has wide applicability for various robotics context detection tasks such as the simultaneous segmentation and detection of text in scenes, 3D classification of terrain types, and grasping and manipulation of objects.

## II. RELATED WORK

The literature on semi and weakly supervised learning is discussed in this section with application to both robotics and existing methods which: 1) generate pseudo labels using a single output modality (e.g. segmentation, CAMs, etc.), and 2) train an ensemble of CNN output modalities to generate pseudo labels.

### A. Semi/Weakly Supervised Learning for Robots

Semi and weakly supervised learning approaches have been used in robotic applications such as: 1) fruit counting for agricultural robots [24], 2) ego-motion estimation [25], 3) object manipulation [26], 4) location detection for topological localization [27], and 5) 3D point cloud segmentation of a scene [28]. Recently, a handful of papers have focused on using semi-supervised learning for semi-automatic labeling of objects from multiple perspectives [29], weakly supervised learning for segmenting terrain [30], [31], and detecting surgical tools [32].

In [29], a semi-supervised method was proposed for generating object labels for industrial robotics. Images of an object were gathered by a camera mounted on a robot arm from multiple viewpoints while tracking the camera pose. The first image was manually labeled with a bounding box label, and all subsequent images were labeled using a relative transform based on the robot's pose when the image was taken.

In [30], a robot was teleoperated through an environment and the robot's footholds were recorded via 2D images taken by an onboard camera and proprioceptive sensors. The footholds in the image frame were labeled manually by terrain type. In [31], images and point cloud data from a 2D camera and LIDAR were recorded while a user drove a vehicle. A traversable path was labeled by registering the path taken into image frames using camera extrinsic parameters and vehicle odometry. Obstacles were detected and labeled in an image using the point cloud data. In both cases, these partially labeled images were used to train a segmentation CNN to predict the labeled path properties.

In [32], a method was proposed for weakly supervised robotic surgical tool detection and localization using image level labels indicating tool presence. During training, an extended spatial pooling layer was applied to the final feature maps of a fully convolutional segmentation network to produce an output classification vector. The vector was trained using a cross entropy loss to predict the presence of a tool. At test time, the position within the feature maps with maximal activation was considered to be the location of a surgical tool.

*B. Single Modality Methods*

In single modality approaches, only one output type is used from the CNN for PSL generation, e.g. per-pixel segmentation. The CNN output is used as a label for self-training to improve the PSLs. Both semi [33], [34] and weakly supervised [15], [19], [35]–[38] methods use this approach to calculate PSLs.

*1) Semi-Supervised Methods:* Semi-supervised single modality methods train a CNN using a combination of fully and partially labeled data to generate pseudo labels for the partially labeled data using the fully labeled data as training labels [33], [34]. For example, in [33], a semi-supervised text segmentation CNN used background-foreground segmentation to train a CNN using synthetic images of text. PSLs were generated automatically using the CNN's segmentation output from the text segments cropped from real images of various environments. These labels were then used to train a text segmentation CNN.

In [34], a pipeline was proposed for training a text detection CNN using semi-supervised character segmentation. First, character segmentation was generated using a synthetic dataset containing character region masks and masks defining the connections between adjacent characters. During training, labels for real images were generated by applying a watershed transform to the CNN's character region output to obtain both the inter-character regions and per-character bounding boxes. The CNN was trained with both synthetic and real data concurrently to update labels for the real images throughout training.

*2) Weakly Supervised Methods:* Weakly supervised single modality methods use only a partially labeled dataset and self-training to generate PSLs. The training targets used in self-training are generated using various different methods including: 1) using CNNs to predict segmentation of an image [15], [19], [35], [36], 2) using CAMs processed by a Conditional Random Field (CRF) [37], or 3) a generative output where a CNN is used to model a joint probability distribution of images and image labels [38].

*a) Segmentation*: In [15], a weakly supervised framework for training segmentation networks was presented. Initial PSLs were generated by using a modified GrabCut [22] method. The masks were combined with objectness proposals from multiscale combinatorial grouping [39] using a union operation. A segmentation network was then trained with these proposals. Labels used for supervision were updated using a set of rules comparing the previous label and segmentation output.

In [19], a weakly supervised object segmentation method consisting of a combination of image level labels and peaks in class response maps (local maxima in a feature map) was proposed. A classification loss was augmented with a peak stimulation function to force the network to focus on discriminative regions. At test time class peaks were detected in the response maps and were refined using peak back propagation to generate an instance segmentation. The mask representing each peak in image space was ranked and then filtered using non-maximum suppression to obtain final object proposals.

In [35], a weakly supervised framework for training segmentation networks using only bounding box annotations was presented. When training the object segmentation network, a graph-based mask refinement technique was used to combine information from the predicted segmentation probabilities, image texture, and ground truth bounding box to update the pseudo labels used as segmentation targets.

In [36], a weakly supervised three-stage framework for the training of a semantic segmentation network using only object labels was proposed. The first stage used a pre-trained unsupervised object segmentation network to generate a coarse initial mask output. The second stage enhanced the generated mask using GrabCut [22] to improve foreground-background segmentation. The third stage involved network training, where, for each training example, a target was generated by processing the network's output using the previous stage.

*b) CAMs*: In [37], a weakly supervised online training procedure was used to train a semantic segmentation CNN based on class labels. The CNN had two parallel output branches, the first was used for CAM and object classification, and the second for per-pixel segmentation. During training, the CAM was filtered using a CRF and converted into a mask containing foreground, background, and unknown information using a heuristic approach to train the segmentation output.

*c) Generative*: In [38], a weakly supervised defect detection procedure using a cyclic generative adversarial CNN and class labels was used. Given both an image with a defect paired and an image without defects the CNN was trained to remove the defect from the image. Similarly, a second CNN was trained to introduce defects into images by using images with defects as ground truth. The outputs were then used as inputs into the opposite generative model and are trained to undo the changes to the image. Segmentation masks were generated by taking the difference between the generated defect free image and the original input and applying heuristic rules.

*C. Ensemble Approaches*

Ensemble approaches combine multiple outputs of a CNN for pseudo label generation, taking advantage of multi-task learning and can therefore improve label quality for self-training [40]. The ensemble combinations include: 1) semi-supervised ensembling of multiple predictions from a single CNN using data distillation [8], 2) weakly supervised ensemble of multiple segmentation maps [41], 3) a weakly supervised ensemble of an attention and saliency map [42], and 4) weakly and semi-supervised ensembles of a CAM and either a saliency map, segmentation mask or intermediate CNN activations [20], [21], [43]–[45].

*a) Data Distillation:* In [8], a pre-trained CNN was used in a semi-supervised data distillation processes to generate labels for unlabeled data. Data distillation involved ensembling predictions using application specific rules from several augmented versions of an input image. Final labels were generated using task specific rules.

*b) Multiple Segmentation Maps:* In [41], a semi-supervised approach that iteratively improves an initial pre-trained segmentation model was presented. Two segmentation models were pre-trained using fully labeled data, the first was trained with bounding boxes and images as input and the second was a self-correction module whose objective was to correct segmentation errors made by the first model. Next, a segmentation CNN taking only the image as input was trained, pseudo labels for the weakly supervised part of the dataset were generated by applying the self-correction network to the CNN's predictions.

*c) Ensembling Saliency and Attention Maps:* In [42], a weakly supervised Expectation Maximization (EM) based method for the generation of segmentation masks using only image labels was presented. An initial estimate for the mask was generated using the per pixel maximum of a class agnostic saliency map and an attention map [46] obtained from pre-existing CNNs via excitation backpropagation. The M-step trained a CNN with a multi-part loss function comparing the posterior and the predicted mask. The E-step was performed by constraining the latent posterior using the image labels, this was used to update target labels at each iteration.

*d) Ensembling of a Combination of Methods:* In [20], a weakly supervised iterative object region erasing approach for object segmentation using class labels was presented. The process first trains a CNN with CAMs to convergence. Using heuristic rules, the CAMs and a saliency map from a pre-trained network were combined to remove aspects of the image discriminative for a particular class. This process was repeated beginning from training until the network no longer converged. Pseudo labels were formed using the removed regions.

In [21], a weakly supervised segmentation CNN trained using class labels was presented. Both intermediate network activations and CAM features were combined to generate a mask label. Intermediate activations taken from layers of the CNN were hand selected based on their apparent discriminative ability. The masks were combined and binarized using heuristic rules and smoothed using a CRF. The predicted mask and class labels were used to train the network's segmentation output.

In [43], a weakly supervised procedure to refine salient object masks generated from an unsupervised method using a CNN trained using class labels was proposed. A CNN was trained using supervision from the original saliency masks and the class label of the image. Then, the original saliency mask, predicted mask, and the top-3 class CAMs were fused using a CRF to produce an updated annotation for each image. The training and label updating process was repeated to generate the final object labels.

In [44], a weakly supervised training procedure was used to train a CNN for salient object segmentation using class labels. A modified CAM layer was used to predict segmentation masks. The network was initially trained for classification, with an L1 penalty on the saliency mask. During training, a saliency map was predicted that was refined using a CRF, which was used to train the network with a bootstrapping loss.

In [45], a CNN with dilated convolution layers was presented for generating pseudo labels in a semi-supervised setting, where only some images had class level labels. The CNN was trained to predict a CAM, and a saliency map. The outputs were combined using a mask merging procedure to update the segmentation target during training with class level labels. When training with fully labeled images, the segmentation output was trained using a per-pixel per-class loss.

*D. Summary of Recent Work*

The aforementioned single modality [15], [19], [33]–[38] and ensemble [20], [21], [41]–[45] methods have shown that pseudo labels can be generated for CNN model training. However, both approaches require large datasets with diverse data to train PSL generation CNNs without overfitting. This may not always be feasible for robots obtaining information from their environments, as datasets of particular environments may be sparse and lack diversity (e.g. available only from one environment). Moreover, semi-supervised techniques require synthetic data or an expert user to fully manually label a large varied subset of data in order to generate labels, which can be a time consuming task [8], [33], [34], [45]. Weakly supervised single modality and ensemble approaches require handcrafted heuristic rules, either for the binarization of masks [15], [20], [37], [38], [43] or for manual network analysis to determine where to gather information from [21]. The approaches that have been designed for robotic applications cannot be transferred to different segmentation tasks, as their training process requires inputs related to their specific problem, e.g. terrain information [29]–[31]. However, weakly supervised methods can generate PSLs using new datasets that do not need to be fully labeled, thus reducing the human time-cost of generating pseudo labels.

In this paper, we present a novel weakly supervised architecture, WeSuperMaDD, which autonomously generates PSLs. The architecture is able to use pre-existing networks that are not trained for a segmentation task to determine PSLs without requiring any additional segmentation training in contrast to prior works. Our approach allows the use of smaller datasets with limited diversity used in robotics as existing CNN models are used as a basis, where the already learned features can be applied to our segmentation task, regardless of their application. To convert these segmentation potential masks (SPMs) into pseudo labels, we introduce a new mask refinement system which incorporates an automated parameter search module that uniquely searches for the smallest PSL. Automating this search eliminates the need for handcrafted heuristic rules to generate PSLs. Therefore, unlike previous methods, we do not require any problem-specific information, thus generalizing our method to wider applications of robot segmentation tasks for obstacle avoidance in cluttered environments, and for grasping and manipulation of objects etc.

III. WEAKLY SUPERVISED MASK DATA DISTILLATION

Our proposed Weakly Supervised Mask Data Distillation architecture, WeSuperMaDD, is presented in Fig. 1 and consists of two sub-systems: 1) Mask Data Generation Sub-System, and 2) Mask Refinement. WeSuperMaDD generates instance level PSLs, $M$, from images of contextual information in an environment. The procedure takes as input a set of CNNs, $\mathcal{F}$, not specifically trained for the task of context segmentation (e.g. trained for object classification, image captioning, etc.), a maximum number of iterations $t_{\max}$, and a dataset containing a set of images, $I$. Each image in the dataset must have a set of bounding quadrilateral labels, and class label $\mathcal{Y}$.

The *Mask Data Generation* sub-system uses the *Image Crop* module to transform each of the context instances within an input image into a standard sized image crop $c \in C$ using the bounding quadrilaterals as reference. A crop, $c$, is then provided to the *Augment* module to generate an augmented set of crops via random image transformations. This set is used in the *Mask Discriminative Region Mining (MDRM)* module to find SPMs, and the predictions made by the CNNs. SPMs are defined as 2D matrices containing the relative segmentation importance of every pixel in the input image.

Both the SPMs and CNN predictions are provided to the *Mask Refinement* sub-system to iteratively generate and refine PSL candidates. The *Segmentation Cost* module determines the distance of each of the predictions with respect to the ground truth. An average SPM is generated via the weighted average of all the SPMs, where each SPM is weighted by its associated cost. A PSL candidate is then generated in the *Pseudo Segmentation Label Generation* module using the initial parameters provided by the *Mask Binary Search* module and the average SPM. The *Mask Attention* module uses this PSL candidate to remove extraneous information from the initial crop, $c$. This focused crop is sent to the *Augment* module to generate a new set of augmented crops. The Ensemble uses the CNNs to generate predictions of the content of each of the crops. The *Segmentation Cost* module then calculates the cost of the predictions. The average prediction cost is used by the *Mask Binary Search* module to update parameters in the *Pseudo Segmentation Label Generation* module. This process is iterated $t_{\max}$ times and at each iteration the current lowest costing PSL candidate is outputted. The two sub-systems are discussed in more detail below.

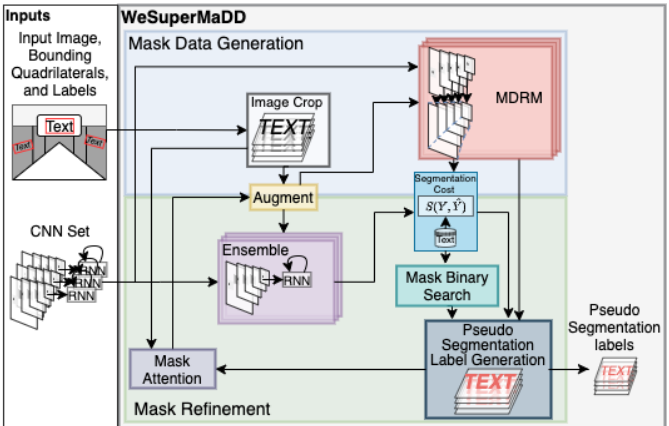

Fig. 1. The WeSuperMaDD architecture.

### A. Mask Data Generation Sub-System

Given an input image and bounding quadrilateral labels representing the outer boundaries of all context instances, the *Mask Data Generation* sub-system produces SPMs, which contain regions in the image that are determined to have discriminative potential. The overall process for the sub-system is as follows:

*1) Image Crop:* The *Image Crop* module takes as input the full image containing contextual information and the bounding quadrilateral labels, and produces images of cropped context instances, Fig. 2. This is achieved using the perspective transformation that changes the context boundary coordinates from the input image into a standard size container, i.e. the input size expected by $\mathcal{F}$. The transform is applied to each of the bounding quadrilaterals to attain the set of crops, $C$. The output cropped context instances are sent to the *Augment* module.

*2) Augment:* The *Augment* module takes as input a single crop, $c$, and generates new samples representing the same context instance with different viewing perspectives and added noise. The objective is to force the CNNs to not rely on a single set of image features to complete the task they were trained for by randomly altering each image. The image augmentation process is adapted from [8], where a single crop, $c$, is randomly augmented to generate a set of crops $C_a$ and inverse transforms, $\mathcal{T}^{-1}$. Augmentations that change the location or shape of context instances must be invertible so predictions can be merged after the inverse is applied. Fig. 3(a) presents sample crops obtained from this process. The augmented crops and their inverse transforms are provided to the *MDRM* module or the *Ensemble* module in the *Mask Refinement* sub-system.

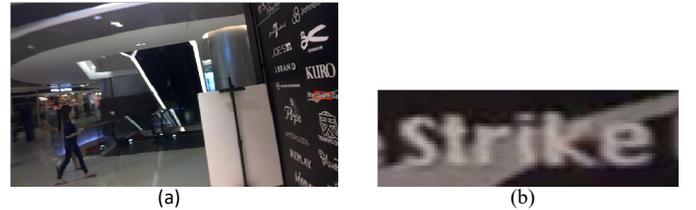

(a) (b)
Fig. 2. Image Crop module: (a) source image from ICDAR-15 dataset with bounding quadrilateral highlighted in red; and (b) a cropped sample.

*3) Mask Discriminative Region Mining (MDRM):* Given the set of augmented crops, the *MDRM* module produces masks representing the discriminative regions of each crop in the form of SPMs. This module uses the features extracted by existing CNNs to generate SPMs for PSL generation. The module begins by performing a forward pass through each of the CNNs in $\mathcal{F}$, and storing their predictions for their respective tasks, $\hat{Y}$, where:

$$\hat{Y} = [\hat{y}_{j,k}] \, \forall \, (c_j, f_k) \in C_a \times \mathcal{F}, \hat{y}_{j,k} = f_k(c_j). \quad (1)$$

We use VisualBackProp (VBP) [47], which examines activations in intermediate CNN layers, to determine the discriminative regions in an input crop to form an SPM. The regions estimate the relative contribution of each of the pixels to the network output $\hat{y}_{j,k}$. This value is used to estimate the relative importance of each pixel for segmentation. VBP has a fast runtime, high visualization quality, and can be easily applied to common network backbones e.g. ResNets [48], VGG [49], etc. Sample SPMs obtained using VBP are shown in Fig. 3(b).

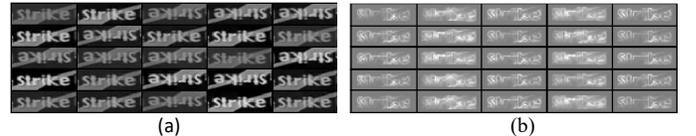

(a) (b)
Fig. 3. Fig. 2(b) after processing by: (a) the *Augment* module; and (b) the SPMs generated by VBP from (a).

After obtaining the SPMs from VBP, the inverse transform corresponding to the crop that was used to generate the mask is applied to attain the set of masks, $\tilde{M}$, where:

$$\tilde{M} = [m_{j,k}] \forall (c_j, f_k) \in C_a \times \mathcal{F}, m_{j,k} = \mathcal{T}_j^{-1}(VBP(c_j, f_k)). \quad (2)$$

and sent to the *Pseudo Segmentation Label Generation* module

in the *Mask Refinement* sub-system. Additionally, the predictions, $\hat{Y}$, are passed to the *Segmentation Cost* module.

*4) Segmentation Cost:* The *Segmentation Cost* module takes as input both a task prediction, $\hat{y}$, and a ground truth label, $y$, and assigns a segmentation accuracy cost to each prediction. In particular, we use a cost function, $s = \mathcal{S}(y, \hat{y})$, to compare the predicted output of a network with the ground truth. The output of this module is either passed to the *Pseudo Segmentation Label Generation* module for mask weighting or the *Mask Binary Search* module to update parameters controlling the creation of PSL candidates, both within the *Mask Refinement* sub-system.

### B. Mask Refinement Sub-System

The mask refinement sub-system takes the SPMs, $\widetilde{M}$, the maximum number of search steps, $t_{\max}$, and the costs, $\mathcal{S}(\hat{Y}, y)$, of each of the predictions made from the augmented data to generate and refine PSL candidate using the set of CNNs, $\mathcal{F}$, where:

$$\mathcal{S}(\hat{Y}, y) = [s_{j,k}] \ \forall \hat{y}_{j,k} \in \hat{Y}, \ s_{j,k} = \mathcal{S}(\hat{y}_{j,k}, y). \quad (3)$$

The process for the *Mask Refinement* sub-system is as follows:

*1) Ensemble:* The *Ensemble* module takes as input a set of crops provided by the *Augment* module, and the CNN set, $\mathcal{F}$, and generates predictions, $\hat{Y}$, made by its members to measure the quality of a PSL candidate. Given the set of crops received from the *Augment* module, a prediction set is created, $\hat{Y} = [\hat{y}_{j,k}]$. The output of this module is sent to the *Segmentation Cost* module to inform the *Mask Binary Search* module.

*2) Mask Binary Search:* The *Mask Binary Search* module uniquely uses the current average prediction cost of the *Ensemble*, $\mathcal{S}(\hat{Y}, y)$, calculated from the cropped images from the *Mask Attention* module. This module autonomously selects the per-image segmentation control parameters. Namely, it updates the current gain, $\alpha^{(t)}$, for the *Pseudo Segmentation Label Generation* module to find the PSL candidate with fewest foreground pixels where the ensemble attains, on average, a cost less than the threshold $s_{\mathbb{1}}$. The initial bounds $u^{(0)} = \max(\widetilde{m})/\text{Otsu}(\widetilde{m})$, and $l^{(0)} = 0$, of the search are set such that the segmentation thresholds can reach the highest and lowest pixel activation of the average mask $\widetilde{m}$. Otsu($\cdot$) [50] is a function that returns a value representing the binarization threshold which minimizes the intra-class variance between the foreground and background classes. The initial gain, $\alpha^{(0)} = 1$, is set such that the search process initialization is unbiased, however, the value can be user selected to incorporate a prior to the search process.

The module tracks and updates the best-known gain $\alpha^*$ and cost $s^*$ according to:

$$\alpha^* \leftarrow \mathbb{1}_{s^*}\alpha^{(t)} + (1 - \mathbb{1}_{s^*})\alpha^*, \quad (4)$$
$$s^* \leftarrow \mathbb{1}_{s^*}s^{(t)} + (1 - \mathbb{1}_{s^*})s^*, \quad (5)$$

where $\mathbb{1}_{s^*} = \mathbb{1}(s^{(t)} < s^*)$, and $\mathbb{1}(\cdot)$ is the indicator function. After finding a PSL candidate that satisfies $s^* < s_{\mathbb{1}}$, the optimization procedure is updated to be:

$$\alpha^* \leftarrow \begin{cases} \alpha^{(t)}, \text{if } s^{(t)} < s_{\mathbb{1}} \wedge \alpha^{(t)} > \alpha^* \\ \alpha^*, \text{else} \end{cases}, \quad (6)$$

in order to store the largest gain adhering to the cost function.

The gain is updated every iteration using:

$$\alpha^{(t+1)} = \mathbb{1}_{s_{\mathbb{1}}}(\alpha^{(t)} + u^{(t)})/2 + (1 - \mathbb{1}_{s_{\mathbb{1}}})(\alpha^{(t)} + l^{(t)})/2, \quad (7)$$

where $\mathbb{1}_{s_{\mathbb{1}}} = \mathbb{1}(s^{(t)} > s_{\mathbb{1}})$. The upper, $u^{(t)}$, and lower, $l^{(t)}$, limits are also updated based on the prediction quality:

$$l^{(t+1)} = (1 - \mathbb{1}_{s_{\mathbb{1}}})\alpha^{(t)} + \mathbb{1}_{s_{\mathbb{1}}}l^{(t)}, \quad (8)$$
$$u^{(t+1)} = \mathbb{1}_{s_{\mathbb{1}}}\alpha^{(t)} + (1 - \mathbb{1}_{s_{\mathbb{1}}})u^{(t)}. \quad (9)$$

The increase or decrease in the gain corresponds to increasing or decreasing the number of foreground pixels in the generated PSL. Fig. 4(a) shows the evolution of PSL candidates generated using the parameters provided by the *Mask Binary Search* module. When the cost given the current PSL candidate is less than $s_{\mathbb{1}}$, the number of foreground pixels in the PSL is decreased by increasing the gain to remove extraneous pixels. When the cost given the current PSL candidate is greater than $s_{\mathbb{1}}$, the gain is decreased to increase the number of foreground pixels in the PSL candidate and incorporate additional image information. The module provides to the *Pseudo Segmentation Label Generation* module either the current calculated gain, $\alpha^{(t+1)}$, or the best gain, $\alpha^*$, when $t = t_{\max}$ number of iterations have elapsed.

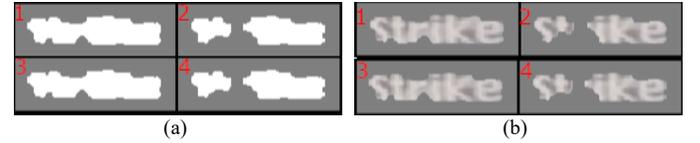

(a) (b)

Fig. 4. Candidate PSLs: (a) obtained by the *Pseudo Segmentation Label* Generation module; and (b) Mask Attention applied to each input crop using the candidate PSLs after each parameter update.

*3) Pseudo Segmentation Label Generation:* The *Pseudo Segmentation Label Generation* module takes the SPMs, $\widetilde{M}$, and determines a candidate PSL. We first ensemble the potential masks to produce a single SPM for each crop using a weighted average using the prediction cost. This reduces the influence of predictions based on non-discriminative regions on the mask ensemble. Given that lower costs indicate a better match we invert the cost and then apply the SoftMax function, $\sigma(\cdot)$, to obtain the relative weighting of each mask:

$$w_{j,k} = \sigma(\max(\mathcal{S}(\hat{Y}, y)) - \mathcal{S}(\hat{Y}, y))_{j,k}. \quad (10)$$

A weighted average is used to combine all of the masks into an average SPM:

$$\widetilde{m}^{(r,s)} = \sum_{j \in [1, |C_a|], k \in [1, |\mathcal{F}|]} w_{j,k} \widetilde{M}_{j,k}^{(r,s)}, \quad (11)$$

where $(r, s)$ is a pixel coordinate. Fig. 5(a) presents a mask representing the weighted average of the SPMs in Fig. 3(b).

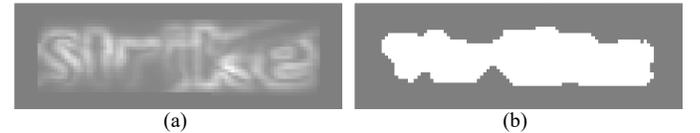

(a) (b)

Fig. 5. (a) The weighted average of the SPMs from the *MDRM* module; and (b) Pseudo label generated after $t_{\max}$ steps based on (a).

We make the assumption that the regions of the image that contain the context instance are within the pixels of the SPM whose activation is greater than some value. We use Otsu's method [49] as the basis for the segmentation thresholds since the mask would be properly segmented if the two classes were clustered in activation. We use the gain parameter, $\alpha^{(t)}$, from the *Mask Binary Search* module to control mask binarization. Therefore, the threshold is set to be:

$$\epsilon^{(t)} = \alpha^{(t)} \cdot \text{Otsu}(\widetilde{m}). \quad (12)$$

We binarize the SPM given the current threshold $\epsilon^{(t)}$, using a threshold operation at each pixel in $\tilde{m}$:

$$\bar{m}^{(t,r,s)} = \mathbb{1}(\tilde{m}^{(r,s)} > \epsilon^{(t)}). \tag{13}$$

We then use the GrabCut algorithm [22] to clean the boundary between the foreground and background. GrabCut uses the input crop image, a Gaussian Mixture Model (GMM), and color to predict the labels of unknown regions of space given an initial foreground and background segmentation. Using the binarized mask, $\tilde{m}^{(t)}$, we generate the unknown region using erosion, $\text{er}(\cdot)$, and dilation, $\text{dl}(\cdot)$, operations. Specifically, we generate our known foreground and background as:

$$m_{\text{fg}}^{(t)} = \text{er}(\bar{m}^{(t)}), \tag{14}$$
$$m_{\text{bg}}^{(t)} = \neg \text{dl}(\bar{m}^{(t)}). \tag{15}$$

Lastly the probable foreground and background regions are:

$$m_{\text{pfg}}^{(t)} = m^{(t)} \wedge \neg m_{\text{fg}}^{(t)}, \tag{16}$$
$$m_{\text{pbg}}^{(t)} = \neg m_{\text{bg}}^{(t)} \wedge \neg m_{\text{fg}}^{(t)}. \tag{17}$$

We apply the GrabCut algorithm using these masks and with an additional set of masks where the probable foreground also includes the foreground mask, selecting the latter if a minimum number of pixels are detected as foreground. We obtain the PSL candidate, $m^{(t)}$, that predicts the class of the pixels in the boundary region shown in Fig. 5(b). The process stops when $t = t_{\max}$, then the mask is provided as the output PSL, $m$. If $t < t_{\max}$, then the PSL candidate is passed to the *Mask Attention* module, continuing the iterative process of PSL generation and evaluation.

*4) Mask Attention:* The *Mask Attention* module is a hard attention operation, that removes pixels of the input crop that are not estimated to be part of the foreground class. In particular, for crop $c^{(t)}$ the PSL is applied as a mask to the input crop as:

$$c^{(t,r,s)} = c^{(r,s)} m^{(t,r,s)} + \mu(1 - m^{(t,r,s)}), \tag{18}$$

thus, leaving only information from discriminative regions by setting the background to a default color $\mu$. Fig. 4(b) shows the evolution of the selected image region using subsequent PSL candidates from the *Pseudo Segmentation Label Generation* module. The masked crops are sent to the *Augment* module to generate samples to evaluate the quality of the PSL candidate.

## IV. WeSuperMaDD Algorithms

The overall architecture is summarized within Algorithm 1 and the *Pseudo Segmentation Label Generation* module (the *PSLGEN* function) is detailed in Algorithm 2:

---
**Algorithm 1** WeSuperMaDD procedure for a single image.

**inputs:**
$t_{\max}$: the number of search iterations, $\alpha^{(0)}$: the initial gain, $\text{fg}_{\min}$: the minimum number of foreground pixels in a PSL, $\mathcal{F}$: the set of pre-trained CNNs, $\mathcal{I}$: image, task labels and bounding quadrilaterals.
**output:**
$(I, \mathcal{Y}, \mathcal{Y}_{\ell}) = \mathcal{I}$ #Image, task label and bounding quadrilateral label
$\mathcal{M}_\mathcal{I} = \emptyset$ #Empty set to hold the new labels for an image
**for** $y \in \mathcal{Y}, g_q \in \mathcal{Y}_q$ **do** #For all context instances in the image
    $h_{c_q}^{g_q} = H(g_q, c_q)$ #Calculate homography
    $c = \text{Interpolate}(I, h_{c_q}^{g_q})$ #Interpolate $I$ to perform crop
    $(C_a, \mathcal{T}^{-1}) = \text{Augment}(c)$ #Generate augmentation samples
    **#MDRM module**
    $\hat{Y} = [\hat{y}_{j,k}] \forall (c_j, f_k) \in C_a \times \mathcal{F}, \hat{y}_{j,k} = f_k(c_j)$ #CNN predictions
    $\tilde{M} = [m_{j,k}] \forall (c_j, f_k) \in C_a \times \mathcal{F}, m_{j,k} = \mathcal{T}_j^{-1}(\text{VBP}(c_j, f_k))$ #VBP masks
    $S = [s_{j,k}] \forall y_{j,k} \in \hat{Y}, s_{j,k} = S(\hat{y}, y)$ #Cost of the predictions
    $\tilde{m} = \sum_{j \in [1,|C_a|], k \in [1,|\mathcal{F}|]} \tilde{M}_{j,k} \cdot \sigma(\max(S) - S)_{j,k}$ #Weighted average mask
    $l^{(0)} = 0$ # Initialize search parameters
    $u^{(0)} = \max(\tilde{m})/\text{Otsu}(\tilde{m})$
    $s^* = -\infty$
    **for** $t \in [0, t_{\max}]$ **do** #Mask binary search loop
        $m^{(t)} = \text{PSLGEN}(c, \alpha^{(t)}, \tilde{m}, \text{fg}_{\min})$
        $c^{(t)} = c \cdot m^{(t)} + \mu \cdot (1 - m^{(t)})$ #Attention
        $(C_a, \_) = \text{Augment}(c^{(t)})$
        $\hat{Y} = [\hat{y}_{j,k}], \hat{y}_{j,k} = f_k(c_j) \forall (c_j, f_k) \in C_a \times \mathcal{F}$ #Ensemble
        $s^{(t)} = \mathbb{E}[S(\hat{Y}, y)]$ #Segmentation cost
        **#Binary search based on average prediction cost**
        $\alpha^{(t+1)} = \mathbb{1}_{s_1}(\alpha^{(t)} + u^{(t)})/2 + (1 - \mathbb{1}_{s_1})(\alpha^{(t)} + l^{(t)})/2$
        $l^{(t+1)} = (1 - \mathbb{1}_{s_1})\alpha^{(t)} + \mathbb{1}_{s_1} l^{(t)}$
        $u^{(t+1)} = \mathbb{1}_{s_1}\alpha^{(t)} + (1 - \mathbb{1}_{s_1})u^{(t)}$
        $s^* \leftarrow \mathbb{1}_{s^*} s^{(t)} + (1 - \mathbb{1}_{s^*}) s^*$
        **if** $s^* < s_1$ **then**
            $\alpha^* \leftarrow \begin{cases} \alpha^{(t)}, \text{if } s^{(t)} < s_1 \wedge \alpha^{(t)} > \alpha^* \\ \alpha^*, \text{else} \end{cases}$
        **else**
            $\alpha^* \leftarrow \mathbb{1}_{s^*}\alpha^{(t)} + (1 - \mathbb{1}_{s^*})\alpha^*$
        **end if**
    **end for**
    $m = \text{PSLGEN}(c, \alpha^*, \tilde{m}, \text{fg}_{\min})$ #Final PSL
    $\mathcal{M}_\mathcal{I} \leftarrow \mathcal{M}_\mathcal{I} \cup \{(m, y, g_q)\}$ #Update the dataset
**end for**
**return** $\mathcal{M}_\mathcal{I}$

---
**Algorithm 2** The *Pseudo Segmentation Label Generation* module (PSLGEN).

**inputs:** $c$: image crop, $\alpha$: current gain, $\tilde{m}$: weighted average of the SPMs, $\text{fg}_{\min}$: minimum number of foreground pixels in a PSL.
**output:**
$\epsilon = \alpha \cdot \text{Otsu}(\tilde{m})$ #Calculate segmentation threshold
$\bar{m} = \mathbb{1}(\tilde{m} > \epsilon)$ #Binarize mask with the threshold
$m_{\text{fg}} = \text{er}(\bar{m})$ #Known foreground
$m_{\text{bg}} = \neg \text{dl}(\bar{m})$ #Known background
$m_{\text{pfg}} = m \wedge \neg m_{\text{fg}}$ #Probable foreground
$m_{\text{pbg}} = \neg m_{\text{bg}} \wedge \neg m_{\text{fg}}$ #Probable background
**#Perform GrabCut optimization to generate PSL candidate *m*, where 0 indicates a mask with only 0's**
$m = \text{GrabCut}(c, \mathbf{0}, m_{\text{bg}}, m_{\text{pfg}} \vee m_{\text{fg}}, m_{\text{pbg}})$
**if** $\sum m^{(r,s)} < \text{fg}_{\min}$ **then** #Check number of foreground pixels
    $m = \text{GrabCut}(c, m_{\text{fg}}, m_{\text{bg}}, m_{\text{pfg}}, m_{\text{pbg}})$
**end if**
**return** $m$

---

## V. Experiments

Our experiments focus on the Optical Character Recognition (OCR) task of the simultaneous detection and segmentation of text in environments due to: 1) its applicability in robotics applications for the detection of text signs to aid for exploration and navigation in unknown cluttered structured environments, and 2) the limitation of existing weakly supervised PSL generation methods as they that are not generalizable to text segmentation since they require the inclusion of additional problem specific NN layers. The weakly supervised generation of text PSLs thus represents a challenging problem not well explored in the existing literature.

The majority of existing publicly available text detection datasets do not have segmentation labels, therefore, to train an instance segmentation CNN for the simultaneous detection and segmentation of text would require segmentation labels to be manually generated by a human expert or by an autonomous

PSL generation method. In these experiments, we investigate the performance of our WeSuperMaDD method in autonomously generating the needed PSLs when only bounding quadrilateral and class label data are available for training.

Herein, we perform three experiments: 1) an Ablation Study, 2) a comparison study of WeSuperMaDD's performance in the generation of PSLs versus other standard methods, and 3) a detailed investigation of instance segmentation of various context images using numerous datasets. The performance metrics used in these experiments are defined as: 1) Precision ($P$), 2) Recall ($R$), and 3) $F_1$ score. Experiments were conducted on a server with a Titan V GPU, an AMD 2990WX CPU, and 128GB of memory.

*A. Ablation Study*

We evaluate the performance of our segmentation method by comparing the class of each pixel of the predicted PSLs and the ground truth masks available in the ICDAR-13 dataset [51]. Within the dataset, pixels overlapping text characters are labeled as foreground, and those not overlapping text characters are labeled as background. We use the ICDAR-13 dataset [51] as it is the only real-world dataset with a per-character text segmentation ground truth with quadrilateral labels. Here we define a positive detection as a predicted PSL pixel matching a ground truth mask pixel (e.g. $m^{(r,s)} = m_{gt}^{(r,s)}$).

*1) Ensemble of SPMs:* To validate the ability of the WeSuperMaDD approach to ensemble SPMs from multiple sources, we generate our CNN set, $\mathcal{F}$, with the following text recognition CNNs which identify the text string contained in an image: 1) a Character Recognition Neural Network (CRNN) [52], and 2) a case-insensitive Thin Plate Spline (TPS) CNN with bidirectional long short-term memory and attention [53]. These CNNs were selected as they use standard structures and are known to be top performers in the text recognition task. To ensemble the SPM predictions we must invert the TPS layer prior to using it in the *Mask Generation* module.

*2) Segmentation Cost Function:* The cost function, $S(\hat{y}, y)$, is modeled using the Edit Distance (ED) between the string predicted by the ensemble and the ground truth string. ED is defined as the minimum number of elementary string operations required to transform one string into the other. We set the threshold, $s_1$ to 1. A cost below this value indicates the CNN was able to identify the text contained within an input image.

*3) Training Datasets:* We train the CRNN with a union of synthetic data from the MJSynth dataset [54] and cropped ground truth regions from the 800,000 images of the SynthText dataset [55], with a combined total of approximately 14 million synthetic English text instances. We also use the union of both the IIIT5k [56] and cropped ground truth text instances from the ICDAR-15 [12] dataset, containing approximately 6,600 text instances from 1,500 images of scene text. For the TPS model, we use an available pre-trained model [57], which was trained using both the MJSynth [54] and SynthText [55] datasets.

*4) Testing Dataset:* Performance was evaluated on the ICDAR-13 dataset [51]The dataset contains 462 images of focused real scene text, with a total of 1,944 text instances. All generated PSLs are resized and compared individually to maintain equal weighting between large and small text instances.

*5) Procedure:* We performed an ablation study to examine the relative importance of hyper-parameters used by WeSuperMaDD with respect to segmentation $F_1$ scores. We used the ICDAR-13 training set and recorded the $F_1$ score obtained using each set of parameters. In particular, we varied: 1) the number of samples generated by the *Augment module,* i.e. $|C_a| = \{1,2,4,8,32\}$, 2) the number of models used in the ensemble, i.e. $|\mathcal{F}| = \{1,2\}$, with only the CRNN ($|\mathcal{F}| = 1$), or both CRNN and TPS ($|\mathcal{F}| = 2$), and 3) the number of binary search iterations i.e. $t_{max} = \{1,2,3,4,5,6\}$.

The results of the ablation study are summarized in Fig. 6. In general, increasing the value of any of the hyper-parameters increases the overall $F_1$ score. As can be seen with one search step either including 2 models or increasing the number of samples to 32 results in increases in $F_1$ score ranging from 12 to 20. Increasing either of these parameters improves PSL generation performance as the activation value of text regions is increased in the SPMs. Thus, the increased activation improves the likelihood that the feature locations will be included in the PSLs.

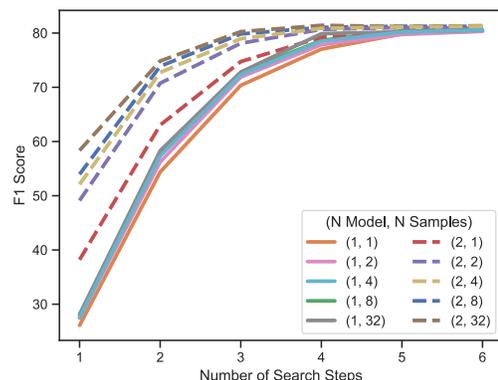

Fig. 6. Ablation of number of models, augmentation samples, and search steps.

When increasing the number of search steps to 6 from 1 step, the $F_1$ score performance improves by 40% when also using 2 models and 32 samples, and 210% when using 1 model and 1 sample. Performance gains are between these two values for all other combinations of the number of models and samples. The increased number of search steps improves the ability of the mask binary search to remove extraneous features from the PSLs as more segmentation parameters are tested. Based on the overall trend, the number of search steps is the most significant determining factor for PSL generation performance.

However, it can be seen that the performance gain decreases as the number of samples increases. For example, only small performance gains are observed with increasing the number of search steps beyond 4 while also increasing the number of samples or models. As can be seen in Fig. 6, when using both 32 samples and 2 models, increasing the number of search steps past 4 to 5 or 6 no longer provides performance gains after reaching a peak $F_1$ score of approximately 81at 4 steps and plateauing. After 4 search steps the CNNs can no longer reduce the size of the segmentation masks while satisfying the cost constraints. Using the knowledge gained from these results, we use 4 steps, 32 samples, and 2 models, to reduce the computational cost of generating PSLs in our next experiments.

*B. Pseudo Segmentation Label Generation Experiments*

We compare the performance of the WeSuperMaDD method against several standard methods. Performance is calculated similarly to the ablation study where the class of each pixel of the PSLs predicted by each method is compared to the ground truth masks available in the ICDAR-13 test dataset [51]. The overall evaluation procedure for individual PSLs is the same as what was presented in experiment *A*.

*1) Methods for Comparison:* We compare our method with PSLs generated using the following common weakly supervised techniques: 1) GrabCut [22], 2) Pyramid [16]*,* and 3) Naive [41]. These methods were chosen since they represent typical gradient free methods that: 1) can be directly applied to an OCR task without the need for training, and 2) have similar label requirements as our proposed method. The methods are applied to generate a PSL for all the context instances of the ICDAR-13 dataset. The GrabCut generation method [22] takes the full image and a bounding box representing the outer edges of a text instance and trains a GMM to segment the background colors from the unknown region inside the bounding quadrilateral which is used as the PSL. The Pyramid generation method generates PSLs using a soft label in the form of a pyramid, the pseudo label peaks at a value of one in the center of a ground truth quadrilateral and decays to a label of zero at the edges [16]. In the Naive generation method, PSLs are generated by labeling the interior of a ground truth quadrilateral text region as foreground [41].

*2) State-of-the-Art Technique:* We additionally compare against the current state-of-the-art semi-supervised Supervision Generation Procedure (SGP) method, whose results are reported in [33]. In [33] a segmentation CNN was trained using text segmentation masks generated using a synthetic dataset generation procedure. The segmentation CNN was then applied to cropped images of real text from the ICDAR-13 dataset in order to generate PSLs.

*3) Segmentation Results:* We generate PSLs for the images in the ICDAR-13 test set using each of the GrabCut, Pyramid, Naive, and our WeSuperMaDD approaches and determine the $F_1$ score for each PSL. The average scores for all four methods are summarized in Table I. A non-parametric Kruskal-Wallis test, $n = 1095$, showed a statistically significant difference in $F_1$ between all methods ($H = 1817, p < 0.001$). Posthoc Dunn tests were conducted between our WeSuperMaDD method and the other three methods and showed that our method has a statistically significant higher $F_1$ with respect to the GrabCut, ($Z = 7.68, p < 0.001$), Pyramid ($Z = 26.22, p < 0.001$), and Naive ($Z = 18.54, p < 0.001$) methods, respectively.

TABLE I
COMPARISON OF SEGMENTATION METHODS ON THE ICDAR-13 TEST SET

| Method | P | R | $F_1$ Score |
|---|---|---|---|
| GrabCut | 52.71 | 72.62 | 57.25 |
| Pyramid | 54.28 | 35.76 | 43.12 |
| Naive | 50.88 | 100.00 | 67.44 |
| WeSuperMaDD (ours) | 73.91 | 88.83 | 80.69 |

Sample PSLs (in red) obtained from the methods are also presented in Fig. 7, where each image is overlaid atop of the ground truth segmentation label. Our WeSuperMaDD approach, Fig. 7(a), is able to find all characters with some small amounts of background information incorporated into the label. The GrabCut result in Fig. 7(b) highlights the problem with using an unsupervised approach, as it only finds a single character rather than considering groups of characters concurrently. On the other hand, the pyramid PSL in Fig. 7(c) is conservative, selecting a fairly small amount of the image as foreground, particularly near the edges of the crop, which significantly reduces the overall score. In the case of the Naive approach, Fig. 7(d), all foreground pixels are included in the PSL, resulting in a 1.0 R, however, significantly more background pixels were labeled as the foreground class giving it the lowest P score (Table I). The inclusion of significant amounts of background pixels is typical to this approach, as it assumes all pixels are part of a foreground object.

*4) Comparison with State-of-the-Art Technique:* Additionally, we compared the $F_1$ scores of the PSLs generated by WeSuperMaDD with the $F_1$ scores from the semi-supervised SGP trained with text segmentation labels published in [33] using images and the same evaluation procedure from the ICDAR-13 test set. The evaluation procedure provides a weighted $F_1$ score for each mask based on the size of the original input image. Our WeSuperMaDD obtained $P$, $R$, and $F_1$ scores of 71.58, 88.15, and 79.00, while the SGP had published results of $P$, $R$, and $F_1$ scores of 89.10, 70.74, and 78.87, respectively. SGP had a higher $P$, while WeSuperMaDD has a higher $R$. However, the overall $F_1$ scores are comparable. We postulate that this performance difference in $P$ and $R$ could be related to the training procedures of the CNNs used within each method. In SGP the training data for the segmentation CNNs can have more background pixels than foreground pixels in their segmentation labels, therefore, the CNN can be biased towards predicting a pixel as background in the presence of uncertainty [58]. The bias results in SGP predicting fewer pixels as foreground, focusing only on obvious true positives as reflected in the results, i.e. higher precision, lower recall. In contrast, WeSuperMaDD is designed to predict pixels as foreground rather than the background as the CNNs incorporated are trained for a text recognition task, where the CNNs must identify all characters in an image to recognize text contained within it. Therefore, the features extracted by the CNNs used in WeSuperMaDD must be from the full character sequence. Consequently, the combination of the *Mask Refinement* and *Mask Binary Search* modules would likely create PSLs that assign more pixels as foreground resulting in a higher recall than precision. This would help to ensure CNNs trained with WeSuperMaDD identify full character sequences. Overall, the average quality measured using the $F_1$ score of the PSLs generated by WeSuperMaDD and the SGP [33], are comparable as a result. Our weakly supervised method, however, has the clear advantage of not requiring any segmentation labels for training and can be thus directly applied to a wide variety of datasets without any human time-effort.

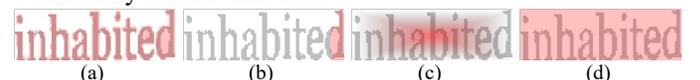

Fig. 7. Sample PSLs overlaid in red on an ICDAR-13 ground truth sample from: (a) WeSuperMaDD; (b) GrabCut; (c) Pyramid; and (d) Naive.

## C. Context Detection and Segmentation Experiments

The objective of the Context Detection and Segmentation Experiments is to evaluate the full text detection and segmentation performance of our WeSuperMaDD architecture.

*1) Training Datasets:* We used the SynthText [55] for pretraining due to its large size and data diversity. The following datasets were then used for fine tuning: 1) ICDAR-13 [51], 2) ICDAR-15 [12], and 3) ICDAR-17 multi-language text [13]. The ICDAR-17 dataset contains 9,000 training/validation images of text from nine languages and six different scripts.

*2) Testing Datasets*: We use: 1) ICDAR-13 [51], 2) ICDAR-15 [12], 3) ICDAR-17 multi-language text [13] datasets, and 4) our own grocery dataset for performance evaluation. The grocery dataset contains 2,226 images that were collected using our mobile interactive Blueberry robot navigating aisles in a real grocery store and focuses on the context detection task of determining real text on aisle signs in the environment. The results are reported using a procedure similar to the ICDAR-15 evaluation procedure [12]. In this case, a positive detection is defined as a predicted quadrilateral with Intersection over Union (IoU) greater than 0.5 with a ground truth quadrilateral, with the additional restriction that a ground truth quadrilateral must be associated with at most one positive detection.

The datasets provide quadrilateral labels of the locations of text instances and text labels for text contained within the images. These datasets provide a variety of real text data in several environments e.g. malls, roads, etc., with varying fonts, scales, and languages, therefore being a useful representation of what a robot may encounter in human-centered environments.

*3) Text Instance Segmentation Model:* WeSuperMaDD uses a modified RetinaNet [59] architecture with a ResNeXt-50 [60] backbone with pyramid levels $P_2$ to $P_6$ for features were used for text detection and segmentation. The selection of a small CNN backbone and a one-stage detection framework reflects the common need of using CNNs for robotics context detection tasks, as small CNNs allow for shorter training time, and faster inference. We modify the RetinaNet architecture to include additional horizontal prior boxes to improve multi-line text detection performance [61]. The bounding quadrilaterals are formed through the prediction of the parameters of a homography matrix that transforms the prior box to a predicted text instance location. The text mask detection branch was shared across all feature maps [9] and uses the rotated RoI-Align module to retrieve features from the backbone [62].

*4) Methods for Comparison:* We compare a text instance segmentation CNN trained with PSLs generated by WeSuperMaDD, referred to as WCNN herein, against a text instance segmentation CNN trained with PSLs generated by the aforementioned Naive approach, referred to as NCNN. The Naive approach was selected as the weakly supervised alternative as it had the highest $F_1$ score after WeSuperMaDD in Table I. Both WCNN and NCNN are based on the aforementioned *Text Instance Segmentation Model.*

*5) State-of the-art-Methods*: Furthermore, we also compare our detection performance against the results reported in the literature for the following methods: 1) Lyu et al. [63], 2) CRPN [64], 3) FOTS [65], 4) Textboxes++ [61], 5) STELA [66], 6) Mask TS [67], and 7) CRAFT [34]. Methods 1-5 are typical text detection CNNs that have been trained using *fully* supervised data, while methods 6 and 7 are *semi-supervised* text detection methods. Furthermore, methods 1-3 and 6 are two-stage methods, and methods 4 and 5 are one-stage methods. CRAFT is a text region segmentation-based method. One-stage methods directly predict bounding quadrilaterals from an image. Two-stage methods extend the one-stage formulation, by including a sub-network that refines the first stage's text proposals. The segmentation method examines a text segmentation output to generate a text bounding quadrilateral. We have selected these methods to investigate the detection performance of WCNN with respect to existing text detection techniques, while uniquely being the only such CNN to provide character level segmentation.

*6) Training Procedure:* Both WCNN and NCNN were trained to output a set of bounding quadrilaterals representing the locations of text within an input image, and a mask highlighting the location of text within this quadrilateral. We used a weighted three-part loss, $L$, containing a bounding box regression loss, a classification loss, and a mask loss, $L_m$:

$$L = (\text{CE}(z_g, z_p) + \alpha_H L_H(\hat{h}_g, \hat{h}_p))/n_{\text{pos}} + L_m, \quad (19)$$

where $n_{\text{pos}}$ is the number of matched prior boxes, and $\text{CE}(z_g, z_p)$ measures the classification loss of the predicted confidence logits $z_p$, and the matching label $z_g$. We rank the prior boxes using the Huber loss, $L_H$, between each prior box and the ground truth quadrilaterals and select the top-$k$, $k = 20$ boxes with lowest loss as positive matches in the label tensor $z_g$. This selection method balances the number of matched prior boxes per ground truth quadrilateral. Out of the remaining prior boxes we select the top-$3n_{\text{pos}}$ with the highest classification loss as the negative examples within $z_g$ [68] to ensure that negative samples are not overrepresented in the loss. Prior boxes matched with ground truth marked as "illegible" within the dataset are not included in the loss calculation.

The $\alpha_H$ hyper-parameter is used to reduce the influence of $L_H(\hat{h}_g, \hat{h}_p)$ on the overall loss [69]. The network is trained to predict the homography that transforms a prior box to a ground truth target. The ground truth homography, $\hat{h}_g$, is the homography of a unit square prior box, $u$, centered at (0,0) to a normalized and centered ground truth quadrilateral, $\hat{g}_g$, to maintain position invariance, $\hat{h}_g = H(u, \hat{g}_g)$. Here $H(\cdot,\cdot)$ represents the solution to the set of linear equations describing the homography between two quadrilaterals. To attain $\hat{g}_g$, we normalize each of the coordinates of $g_q$ as:

$$\hat{g}_g = \left(\frac{g_{q_{x_i}} - b_{c_x}}{b_w \sigma_{b_w}^2}, \frac{g_{q_{y_i}} - b_{c_y}}{b_h \sigma_{b_h}^2}\right), \forall i \in [1,4], \quad (20)$$

using the center $(b_{cx}, b_{cy})$, side lengths $(b_w, b_h)$, and length variances $(\sigma_{b_w}^2, \sigma_{b_h}^2)$ of the matched prior boxes.

Instance segmentation is trained using the mask loss [9]:

$$L_m = \frac{1}{k_s}\sum_{k=1}^{k_s} CE(M_g^{(k)}, M_p^{(k)}) / |M_g|, \quad (21)$$

where $CE(M_g^{(k)}, M_p^{(k)})$ is a cross entropy loss between each predicted mask pixel and the PSLs. The loss function is applied to the masks generated from the top-$k_s$, $k_s = 50$ bounding quadrilateral predictions [62].

Datasets with similar image distributions are grouped for fine-tuning and testing similar to [34]. When testing on ICDAR-13, and ICDAR-17 we fine-tune using their training sets. When testing on the ICDAR-15, and our own real grocery store dataset we fine tune using the ICDAR-15 training set. We apply random color, perspective, rescaling, and cropping perturbations during training and PSL generation. Before training either WCNN or NCNN we generate PSLs by applying WeSuperMaDD or the Naive approach for each image in the dataset.

*7) Testing Procedure:* During testing, we resize the longer edges of images from ICDAR-13, 15, 17, to 960, 1,920, and 1,600, respectively. For our own grocery dataset, the image size of 1280x720 pixels is used to compare both the WCNN and NCNN to measure their detection and segmentation performance in a real robot context detection task.

As the ICDAR-15, 17 and our grocery datasets do not provide instance segmentation labels, we compared the ability to localize text with the masks generated by WCNN and NCNN using a Proposal Refinement Procedure (PRP) we developed. In particular, we generate text detection quadrilateral proposals from segmentation masks using the minimum rotated rectangle surrounding the mask. The procedure begins by converting a quadrilateral text region proposal into a rotated rectangle and using the *Mask Generation module* to segment the text contained within it. We iteratively move each edge of the region proposal outwards until at least *n* of the pixels on that edge are foreground. After a specified number of iterations, we generate the final proposal using the minimum rotated rectangle surrounding the segmentation mask. The procedure tests whether the bounding rotated rectangle proposal can be recreated using the predicted mask. Given a predicted quadrilateral, if the quadrilateral obtained from the procedure is larger than the corresponding ground truth quadrilateral, then background information was incorporated into the mask. If the quadrilateral is smaller, a part of the text instance was not detected. This process examines the ability of the CNN to 1) find the characters within a region, 2) to differentiate individual text instances, and 3) to detect the ability of the CNN to localize text. An example of this process applied to both methods is shown in Fig. 8, where initial masks, Figs. 8(a) and 8(c), are refined using the described procedure to grow the initial mask proposals as shown in Figs. 8(b) and 8(d) respectively.

*8) Text Detection Results:* The text detection results for WCNN and NCNN are presented in Table II. For each of the ICDAR-13, 15, 17, and grocery datasets, we observe an $F_1$ score ranging from 68.85 to 89.56 for WCNN and an $F_1$ score ranging from 68.56 to 89.44 for NCNN, respectively The results show WCNN has only a slight improvement in $F_1$ scores between 0.12 and 1.50 higher than those achieved by NCNN for each of the datasets. The slight performance improvement is likely due to WCNN learning more robust features for instance segmentation and is related to the same detection labels being used to train either CNN. Since the WeSuperMaDD PSLs provide more accurate localization information for text, they have better text segmentation output quality for localization than NCNN, as is explored in the following section.

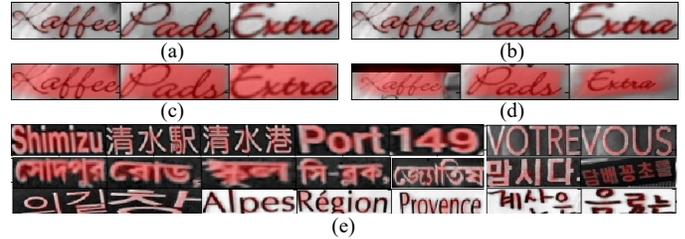

Fig. 8. Sample crops from ICDAR-13 dataset overlaid with: (a) masks generated from WCNN; (b) masks generated from NCNN; (c) masks from (a) after refinement; (d) masks generated from (b) after refinement; and (e) randomly selected segmentation samples generated by WCNN overlaid on ICDAR-17 validation images.

*9) Text Segmentation Results:* We also compare the segmentation performance of both WCNN and NCNN based on their final $F_1$ scores after applying the PRP (Section V.C.4). Since only WCNN and NCNN provide a text instance segmentation output, we, therefore, cannot compare against the existing state-of-the-art techniques using this procedure. The results are summarized in Table II. For each of the ICDAR-13, 15, 17, and grocery datasets we observed an $F_1$ score ranging from 67.10 to 85.24 for WCNN and an $F_1$ score ranging from 41.76 to 71.03 for NCNN, respectively. We note that WeSuperMaDD PSLs provide a gain of between 42% and 83% in terms of $F_1$ when compared to the Naive PSLs. The low score attained by NCNN can be attributed to the label quality of the naive PSLs. Namely, since the Naive method mislabels background pixels near the edges of ground truth bounding quadrilaterals, the edges of text segmentation masks will also be incorrect. In contrast, WeSuperMaDD PSLs are designed to only label character font as foreground, reducing the ambiguity of the location of the edge of a text instance thereby improving the $F_1$ score.

We further conducted a qualitative comparison of WCNN and NCNN. Figs. 8(a) and (b) show a segmentation proposal from WCNN before and after mask refinement. We note that there is minimal change to the bounding quadrilateral proposals since the initial segmentations were within the boundary of the proposals. This contrasts to the masks generated by the NCNN which incorporates background pixels surrounding text as seen in Fig. 8(c) and results in the refinement procedure increasing the size of the bounding quadrilateral in 8(d) to the point that they no longer satisfy the IoU criteria for a positive detection. Therefore, WCNN is able to produce instance segmentation masks that are more informative of the location and size of text than those generated by NCNN. WeSuperMaDD is able to generate PSLs that only segment characters within the text bounding boxes, providing higher precision guidance for WCNN to localize text boundaries.

We provide several examples of text instance segmentation masks generated by WCNN on the ICDAR-17 validation set within Fig. 8(e) for further qualitative analysis. It can be seen that WCNN provides segmentation proposals (red) for multiple scripts and languages that overlap the text in the source image while incorporating minimal background pixels. It is interesting to note that this is despite using CNNs that have only been trained on English text and Latin script to generate the PSLs. The ability of WCNN to detect and segment important features in text in varying languages and scripts highlights the generalizability of the WeSuperMaDD pseudo label generation process.

TABLE II
COMPARISON OF DETECTION AND SEGNEMNTATION METHODS ON THE TEST DATASETS. * INDICATES A ONE-STAGE DETECTOR, ** INDICATES A MULTI-STAGE † INDICATES A SEGMENTATION-BASED METHOD.

| Method \ Dataset | ICDAR-13 | | | ICDAR-15 | | | ICDAR-17 | | | Grocery | | |
|---|---|---|---|---|---|---|---|---|---|---|---|---|
| | P | R | $F_1$ | P | R | $F_1$ | P | R | $F_1$ | P | R | $F_1$ |
| Detection Results | | | | | | | | | | | | |
| WCNN* | 90.72 | 88.44 | 89.56 | 91.05 | 84.69 | 87.75 | 78.94 | 61.05 | 68.85 | 82.34 | 92.89 | 87.3 |
| NCNN* | 90.05 | 88.84 | 89.44 | 89.36 | 83.34 | 86.25 | 75.77 | 62.6 | 68.56 | 82.64 | 91.94 | 87.04 |
| Segmentation Results | | | | | | | | | | | | |
| WCNN* | 85.41 | 81.97 | 83.65 | 81.42 | 88.21 | 84.68 | 79.64 | 57.98 | 67.1 | 80.21 | 90.94 | 85.24 |
| NCNN* | 70.79 | 71.26 | 71.03 | 66.78 | 66.97 | 66.88 | 45.76 | 50.62 | 41.76 | 49.7 | 77.86 | 60.68 |
| Semi-Supervised Detection Results | | | | | | | | | | | | |
| CRAFT† [34] | 97.40 | 93.10 | 95.20 | 89.80 | 84.30 | 86.90 | 80.60 | 68.20 | 73.90 | - | - | - |
| Mask TS** [67] | 95.00 | 88.60 | 91.70 | 91.60 | 81.00 | 86.00 | - | - | - | - | - | - |
| One/Two-stage Fully Supervised Detection Results | | | | | | | | | | | | |
| Lyu et al.** [63] | 93.3 | 79.40 | 85.80 | 94.10 | 70.70 | 80.70 | 83.80 | 55.60 | 66.80 | - | - | - |
| CRPN** [64] | 92.10 | 84.00 | 87.90 | 88.70 | 80.70 | 84.50 | - | - | - | - | - | - |
| FOTS** [65] | - | - | 88.30 | 91.00 | 85.17 | 87.99 | 80.95 | 57.51 | 67.25 | - | - | - |
| Textboxes++* [61] | 88.00 | 74.00 | 81.00 | 87.2 | 76.7 | 81.7 | - | - | - | - | - | - |
| STELA* [66] | 93.30 | 85.10 | 89.00 | 88.70 | 78.60 | 84.50 | 78.70 | 65.50 | 71.50 | - | - | - |

*10) Comparison to State-of-the-Art Detection Techniques*: We compared the performance of our one-stage WCNN to seven other approaches presented in the literature with the ICDAR-13, 15, and 17 datasets, Table II. Our WCNN outperforms all existing one-stage detectors on the ICDAR-13, 15 datasets with an $F_1$ score increase of 0.56 and 3.25. However, on the ICDAR-17 dataset, it had an $F_1$ score 2.85 lower than STELA. We postulate that the reason that WCNN outperformed the other one-stage detectors on the ICDAR-13 and 15 datasets was due to our unique use of the text instance segmentation training. This segmentation training helps the network distinguish text regions from background regions due to the inherent per-pixel detail of PSLs. In contrast, the other one-stage methods only use bounding quadrilaterals during training which only provide broad guidance of what image regions contain text. We hypothesize that the reason why the STELA had better performance on the ICDAR-17 dataset was due to it directly learning the shape and location of the prior boxes used during the detection task via regression which can improve the number of prior boxes that satisfy the ground truth bounding quadrilateral matching criteria. This is important for large sized bounding quadrilaterals since they typically only match with one prior box and are thereby underrepresented in training resulting in lower detection performance. This was evident for the ICDAR-17 dataset since it contains several text instances that are very large in area.

When compared to the remaining two-stage and segmentation methods on the ICDAR-13, 15, and 17 datasets, WCNN had slightly lower $F_1$ scores of 5.64, 0.24, and 5.05 on each dataset, respectively. This is not unexpected given that two-stage models are known to outperform one-stage models [68], especially in the case of semi-supervised methods that are trained using a fully labeled subset of data. However, the clear advantage of our method with respect to all these two-stage methods is that WCNN can provide a segmentation output without the need for labeled data and with more accurate localization information, as it provides per-pixel segmentation, while other methods rely purely on using bounding quadrilaterals for text localization. Semi-supervised methods cannot be generalized to all datasets due to the need for a fully labeled subset of the dataset. Overall it can be seen that WCNN has comparable performance to existing detection models since it outperforms all of the alternative methods on **at least** one of the three datasets tested.

## VI. CONCLUSIONS

In this paper, we present the novel WeSuperMaDD method for the weakly supervised generation of PSLs of contextual information datasets collected in various environments. The novelty of the method is that it can autonomously generate PSLs using pre-existing CNNs not specifically trained for the segmentation task. Our method uses learned image features to directly generate these labels for small and non-diverse datasets typically present in robotic environments including grocery stores, malls, roads, etc. A new mask refinement system is introduced to find the PSL with fewest foreground pixels that satisfies constraints as measured by a cost function. The mask refinement system removes the need for handcrafted heuristic rules typically needed by existing PSL generation methods. Experiments validating the performance of WeSuperMaDD were conducted using OCR datasets due to the wide applicability of detecting scene text for robotic applications such as navigating unknown environments, annotating maps, etc. The datasets contained images of text with varying languages, scripts, and scales obtained from various environments including parks, border crossings, and shopping malls The experiments showed that PSLs generated by WeSuperMaDD had: 1) significantly higher $F_1$ scores for segmentation when compared to other weakly supervised methods including: GrabCut, Pyramid, and Naive, and 2) comparable $F_1$ scores to the current state-of-the-art semi-supervised PSL generation method. We further validated our overall architecture for instance segmentation and detection of real text in varying indoor and outdoor environments. We found our CNN trained with WeSuperMaDD PSLs has a higher segmentation $F_1$ scores than a context segmentation CNN trained with Naive PSLs and higher detection $F_1$ scores than existing state-of-the-art one-stage models. To further increase the applicability of our method to a variety of robotics tasks, future work will consider extending our approach for datasets with only bounding boxes or class labels to further reduce the burden of labeling.